# Classification for Big Dataset of Bioacoustic Signals Based on Human Scoring System and Artificial Neural Network

**Mohammad Pourhomayoun**                                                                MP749@CORNELL.EDU
**Peter J. Dugan**                                                                        PJD78@CORNELL.EDU
**Marian Popescu**                                                                         CP478@CORNELL.EDU
Bioacoustics Research Program (BRP), Cornell University, Ithaca, NY, USA, 14850

**Denise Risch**                                                                      DENISE.RISCH@NOAA.GOV
Northeast Fisheries Science Center, Woods Hole, MA, USA, 02543

**Harold W. Lewis III**                                                                HLEWIS@BINGHAMTON.EDU
Department of Systems Science and Industrial Engineering, Binghamton University, NY, USA, 13850

**Christopher W. Clark**                                                                   CWC2@CORNELL.EDU
Bioacoustics Research Program (BRP), Cornell University, Ithaca, NY, USA, 14850

## Abstract

In this paper, we propose a method to improve sound classification performance by combining signal features, derived from the time-frequency spectrogram, with human perception. The method presented herein exploits an artificial neural network (ANN) and learns the signal features based on the human perception knowledge. The proposed method is applied to a large acoustic dataset containing 24 months of nearly continuous recordings. The results show a significant improvement in performance of the detection-classification system; yielding as much as 20% improvement in true positive rate for a given false positive rate.

## 1. Introduction and Background

Passive acoustic monitoring is one of the primary and popular methods used to help scientists investigate and understand animal behavioral patterns (Potter, Mellinger, & Clark, 1994). The acoustic modality is particularly appropriate for marine mammals, because all of those studied are known to produce sounds for foraging, navigation or communication. Furthermore, acoustic monitoring methods are not subject to visual sighting limitations imposed by weather, daylight and ocean environmental conditions (Norris, Oswald, & Sousa-Lima, 2010). In passive acoustic monitoring, fixed acoustic sensor systems record the underwater sounds. These systems collect huge amounts of acoustic data. Exploring the data can be done by the human. The inspection however, is inefficient and slow. Instead, advanced computer algorithms have been designed to identify various animal sounds which tend to augment the humans ability, making this process more efficient for data analysis.

For decades, scientists have been actively recording and archiving marine bioacoustic data, and have devoted significant effort at designing automated algorithms for processing these data for sounds of interest. Processing the data poses many challenges including highly variable ambient noise conditions and a host of biological and anthropogenic sound sources.

The process of bioacoustic signal identification usually includes three main stages; signal detection, feature extraction, and classification. One of the most widely used detectors for acoustic signals is the Energy Detector (Ichikawa et al., 2006; Jarvis, DiMarzio, Morrissey, & Morretti, 2006; Ura et al., 2004). However, it suffers greatly when the signal to noise ratio (SNR) is low. One solution to address this problem is to denoise the incoming data before the detection step (Datta & Sturtivant, 2002; Gillespie et al., 2008; Ichikawa et al., 2006; Niezrecki, Phillips, Meyer, & Beusse, 2003; Popescu et al. 2013). Adjusting thresholds remains a common issue when balancing between false positive (false alarm) and false negative (missed detection) errors. For example, with endangered animals, populations are typically very low and the detection requirement aims to minimize the false negative rate. For animals that are relatively abundant, reducing the false positive rate may be a more optimal detection requirement (Dugan, Rice,

Urazghildiiev, & Clark, 2010). The detection threshold offers the ability to vary these conditions.

The second stage in the process of bioacoustic signal identification is feature extraction from the detected sound events. These features can then be used as input for the classifier. Classification is typically the final stage for a typical computer algorithm. Since classification is highly dependent on prior information, provided by detection and feature extraction, it is often the most critical step for marine mammal identification. Similar sounds produced by various species, overlapping vocalization and interfering noise are common reasons that make marine mammal acoustic classification difficult. However, several effective and promising methods are being developed and used by researcher in addressing this issue (Afifi & Clark, 1996; L. A. Clark & Pregibon, 1991; Deecke, Ford, & Spong, 1999; Mazhar, Ura, & Bahl, 2007; Mellinger, 2004, Mellinger, D.K; Murray, Mercado, & Roitblat, 1997; Norris et al., 2010; Dugan, Rice, Urazghildiiev, & Clark, 2010). Fig. 1 shows a sample time-frequency spectrogram including four different types of acoustic events (i.e. objects) overlapping with each other in a short period of time.

Experiments show that for small datasets, researchers have been successful at finding combinations of detector-classifier pairs that produce satisfactory results. However, when acoustically sampling over long time periods (months to years) and geographical areas ($10^5$ km$^2$), artifacts in the sound environment tend to color the spectrum and make automatic recognition less successful (especially increases in rates of false positive errors) (Clark et al., 2010; Dugan et al., 2010).

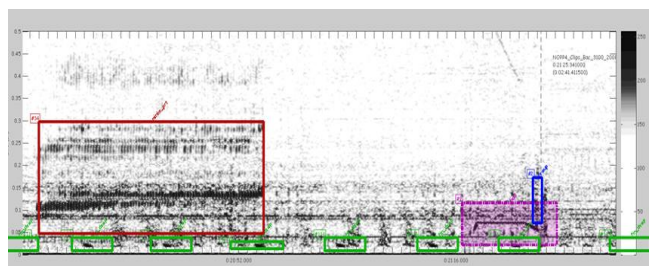

Figure 1. Example spectrogram including a minke whale pulse train song (red), fin whale song notes (green), a right whale up-call (blue) and hard disk drive noise (pink).

In this paper, we propose a technique for improving the algorithm recognition results for use with large datasets. The technique herein is particularly useful when human operators want to visualize seasonal activity or diel patterns over long time periods. This approach uses human knowledge along with information from the recognition system, such as signal features or recognition parameters, to improve the overall performance of the detection-classification system. The proposed method is used as a post-processing stage for existing algorithms, and works by classifying based on human knowledge along with recognition parameters. For this work, classification is done through the use of the Artificial Neural Network (ANN). Combining human perception along with recognition parameters improves the performance for seasonal activity of a marine mammal species commonly known as the minke whale (*Balaenoptera Acutorostrata*). We refer to our algorithm as the human knowledge-ANN, or HK-ANN. Since minke whale populations are broadly distributed throughout an ocean basin, for this work we are not interested in finding every call. Instead, producing a quantitatively based, geographic map of seasonal occurrence and distribution with minimal human involvement is the primary biological goal. For this work, we will show how the HK-ANN is first trained and tested on a smaller dataset using labeled events and quality scores. By allowing the neural net to utilize the human knowledge during the training, the HK-ANN is then applied to 24 months of nearly continuous sounds (from 2008 to 2010) recorded in the Stellwagen Bank National Marine Sanctuary (SBNMS), Massachusetts Bay, United States (Morano et al., 2010). The results show a significant improvement in performance of the detection-classification system in identifying minke whale sounds. The outcome also shows a remarkable improvement in eliminating the false alarm errors during the 24-month period by using the proposed method. Results are described by comparing the diel activity pattern for the 24-month case. We will show that the HK-ANN achieves a true temporal/seasonal minke whale song distribution pattern, with minimal false positive errors.

## 2. Bioacoustic Signal Processing and Classification

### 2.1 ANN Classifier for Bioacoustic Signals

The Artificial Neural Network (ANN) is an effective method for acoustic signal classification (Mellinger, 2004; Murray et al., 1997; Potter et al., 1994) (Deecke et al., 1999; Dugan et al., 2010). There are several reasons that make ANN a promising method. First, ANN is a non-linear estimator. Thus, it can be well-suited for noisy inputs with arbitrary distributions, especially when the interfering noise is not statistically independent of the desired signal (Potter et al., 1994). Second, ANN is an adaptive classifier. Feed-forward ANN can be trained by interactive methods that adjust the weighing matrix to minimize the cost function and to guarantee achieving an (at least a local) optimal weighing network model (Potter et al., 1994). Moreover, ANN can take metrics from a wide range of acoustic representations (e.g. spectrogram, waveform, frequency contour) as input. This flexibility helps in a variety of applications and supports designing different classifier topologies for several different signal types. For example, Deecke *et al.* (1999) used a standard back-propagation trained ANN to classify killer whale dialects to nine different categories by using the extracted pulse-rate contours of killer whale signals as input to an ANN. Potter *et al.* (1994) used a feed-forward ANN to distinguish bowhead whale song endnotes from interfering noises, and they used the signal spectrogram as the input of an ANN.

There are several types of ANNs that can be applied for the purpose of bioacoustic signal classification. Feed-forward networks are more preferred for our purpose because they have less complexity and a relatively lower number of neurons and connections when compared to other networks (Potter et al., 1994). However, feed-forward networks usually need a larger training set for the learning phase (Potter et al., 1994). In our case, this is not a problem because we have a large amount of data available for training. Thus, we preferred to choose a standard feed-forward, back-propagation trained network for our marine mammal sound post-classification task. A feed-forward neural network usually includes at least one hidden layer. The output of each hidden layer is the non-linear function of the linear combination of its input data coming from the previous layer. The coefficients in each combination (called weights) are adaptive parameters adjusted during the training step. The $i^{th}$ output of the first hidden layer is calculated as the sum of weights and a bias term,

$$y_i^{(1)} = f(\sum_{n=1}^{N} w_{in}^{(1)} x_n + b_i^{(1)}) \qquad (1)$$

where $f(.)$ is the non-linear function, $x_n$ is the $n^{th}$ input element, $w_{in}^{(1)}$ is the weight element of the first layer weight matrix and $b_i^{(1)}$ is the bias. Similarly, for the second layer we have,

$$y_i^{(2)} = g\left(\sum_{m=1}^{M} w_{im}^{(2)} f(\sum_{n=1}^{N} w_{mn}^{(1)} x_n + b_m^{(1)}) + b_i^{(2)}\right) \qquad (2)$$

where $g(.)$ is the nonlinear function again, and $w_{in}^{(2)}$ is the weight element of the second layer. The network may be trained in a variety of ways including simple gradient descent, where we represent the collection of synaptic weights as the vector $\bar{w}$, the gradient of the error function as $\bar{g}$, and a learning rate $\rho$, iteratively calculating $\bar{w} = \bar{w} - \rho \cdot \bar{g}$ until an appropriate level of convergence is attained.

**2.2 ANN Post Classification Based On Human Scoring**

Minke whales are known to sing, and singing occurs seasonally in different regions of an ocean. Minke songs consist of 40-60 sec sequences of short duration (40-60 msec), broadband (ca. 100-1400 Hz) pulses, referred to as a pulse train (as shown in Figure 2-(a)). To achieve a higher performance in identifying the minke whale calls in a large dataset, we designed and implemented a post-classifier process based on human expertise, and applied it on the output of an existing detection-classification system. Note that the goal of this paper is not to compare the performance of ANN-based classification to other types of classifiers. Instead, we consider the existing detection-classification system as a black box, and we aim to improve its overall performance using human-knowledge post-processing approach.

In the detection stage, we used a simple energy approach (Datta & Sturtivant, 2002; Gillespie et al., 2008; Ichikawa et al., 2006; Jarvis et al., 2006; Niezrecki et al., 2003; Ura et al., 2004) in addition to color compression and image processing methods (Witten, 2011) to detect acoustic events in the time-frequency domain. Each event can be either a minke pulse train or other sounds, which include ambient noise, anthropogenic noise, or the acoustics of other marine mammals.

Afterwards, a feature set (feature vector) for each event was extracted from the original time-domain signal as well as the signal spectrogram. We used the signal feature vector as the ANN's input and consider a score assigned to each event as the ANN's expected output. The input feature vector includes 18 features such as event duration, event minimum and maximum frequencies, number of pulses in the pulse train, average bandwidth, center frequency, equivalent continuous sound pressure level ($L_{eq}$), mean, mode, maximum and minimum of the pulse duration and pulse intervals, as well as SNR with respect to $5^{th}$, $10^{th}$, $20^{th}$ and $25^{th}$ percentile of the signal.

For this work, we used a heuristic approach to derive the training parameters. We used a training dataset containing 2625 events. Note that all of these events have been already identified as minke pulse trains (i.e. songs) by the existing detection-classification system; however, they include a large amount of false positive errors. The main goal of using the post-classifier is to improve the performance by eliminating these false positives by exploiting a combination of human expertise and machine-learning techniques.

After selecting the training set, expert biologists assigned a score to each one of the 2625 events. Scores were assigned by evaluating the spectrogram of the sound signal. The scores varied from 0 to 4 and were defined as following; 0: *Not target species,* 1: *Unsure of target species,* 2: *Faint target species,* 3: *Mediocre target species,* 4: *Strong target species.* According to the scores assigned by the human expert, only 981 of the events were likely to be minke pulse trains with scores of greater than or equal to 3. Figure 2 shows the spectrogram of four sample events scored from 1 to 4 by an expert biologist. Quality scores based on human intuition were added to the standard training set as shown in (3),

$$TV_{i,j}^{HK} = \left\langle FV_{i, j=1 \cdots 18}, S_{k \in [0,4]} \right\rangle \qquad (3)$$

where $FV_{i,j}$ is the $j^{th}$ feature, ranging from 1 to 18 for the $i^{th}$ object in the training set. The output class is given by scores $S_{k \in [0,4]}$ as mentioned above.

It was discovered that an acceptable convergence was obtained by using three hidden layers. Hidden layers used a *sigmoid* activation function, and the output layer applied *softmax* activation normalized to the interval shown in equation (3). The ANN was initialized with random weights, and training was accomplished by correcting the weights iteratively using backpropagation rule of steepest descents and minimizing the mean squared error. After around two hundred training iterations, the mean square error stabilized to less than 0.01, which was acceptable for our purpose.

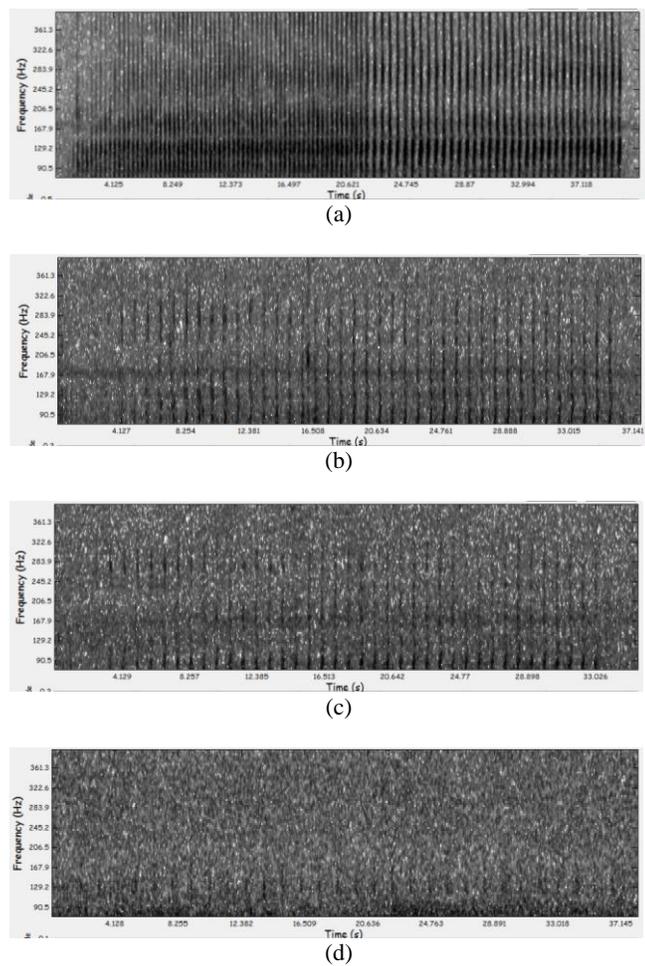

Figure 2. Spectrogram of the 4 sample events scored by human expert. (a) score = 4, (b) score = 3, (c) score = 2, (d) score = 1.

## 3. Results

The designed HK-ANN was tested on a big dataset including 24 months of nearly continuous data recorded in the Stellwagen Bank National Marine Sanctuary (SBNMS), Massachusetts Bay, United States. The testing dataset contains 41560 sound events detected as minke pulse trains by the primary detector. To evaluate the performance of the proposed method, we asked the human expert to mark the minke pulse trains in the entire dataset. We then compared this marked dataset against HK-ANN classifier output using a temporal/seasonal pattern (diel pattern).

The diel pattern shows the distribution of the bioacoustic events in a date-time plane, as shown in Figure 3. The shaded area represents nighttime, while the white area shows day time (the length of day time changes along the calendar). Horizontal greyish strips illustrate periods when the recording sensors were being recovered and redeployed , so no data were available for those time periods. Figure 3-(a) shows the classification results of the primary *decision tree classifier* using the same 18-feature set. However, given our prior knowledge on minke whale seasonal distribution in the sampling area, we expected a large portion of these primary detections to be false alarms. Figure 3-(b) demonstrates the true results identified by the human expert. Comparing figures (a) and (b), we can observe a significant rate of false positives in the classification stage. Figure 4-(c) shows the proposed HK-ANN classification results when we consider the output events, with a score equal to 4, as minke vocalization. Comparing this figure to traditional decision tree classifier (figure (a)) and the truth set (figure (b)) shows a big improvement in the identification of minke vocalization and a reduction in false positives. Figure 4-(d) represents the proposed HK-ANN classification results when we consider the output events, with a minke vocalization score greater than or equal to 3.

Biologists are usually interested in investigating the behavior of the animals during a specific time period. Marine mammals typically exhibit a seasonal pattern, and thus one of the most important parameter showing the performance of a marine mammal detection/classification method is to achieve the correct temporal/seasonal animal distribution. As we see in figure 3, the proposed method is able to eliminate a big portion of the false alarms and achieve a fairly accurate animal temporal/seasonal distribution pattern.

Figure 4 shows the Receiver Operating Curve (ROC) performance of the various common classifiers including Bayesian Network Classifier (Duda, Hart, & Stork, 2001; Witten, 2011), Grafted Decision Tree Classifier (Webb, 1999), and Classification/Regression Tree Classifier (Breiman, Friedman, Olshen, & Stone, 1984), on a dataset containing 4474 highly noisy sound samples. The blue curve shows the performance of the system after applying the proposed post-classifier HK-ANN on the output of above classifiers for the same dataset. We can see a significant improvement in overall performance (especially at low False Positive Rates), by using the proposed method. For example, at a FPR of 6% we have an improvement in TPR of approximately 20%. This improvement can help biologists make better informed and more accurate decisions about marine mammal seasonal occurrences and distributions.

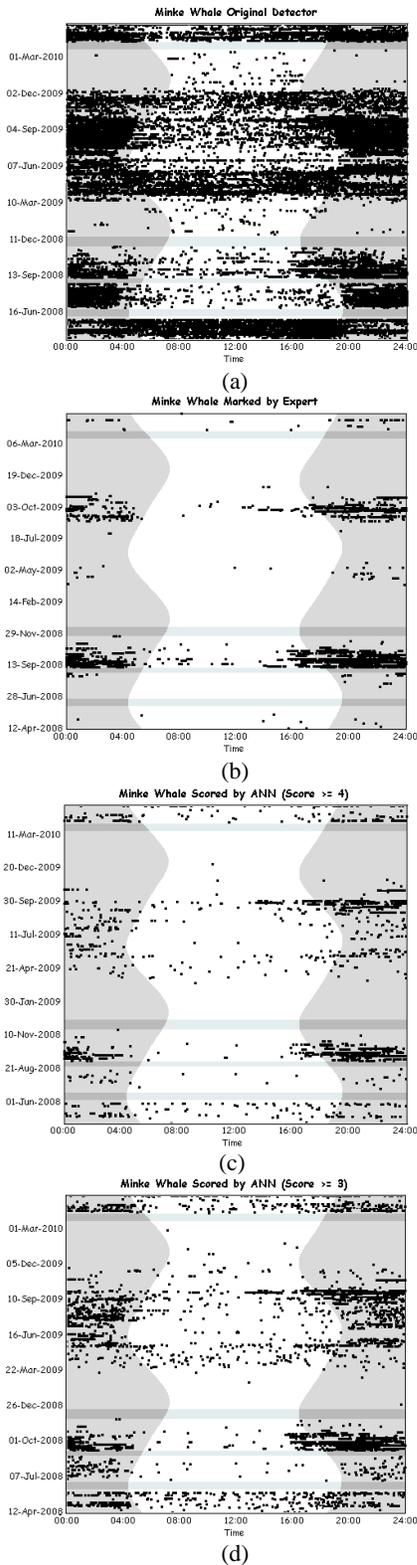

Figure 3. Date versus time diel patterns for test dataset. (a): Original detection/classification by existing decision tree classifier. (b): True detections by human expert. (c): Detection by ANN with score=4. (d): Detection by ANN with score > 3.

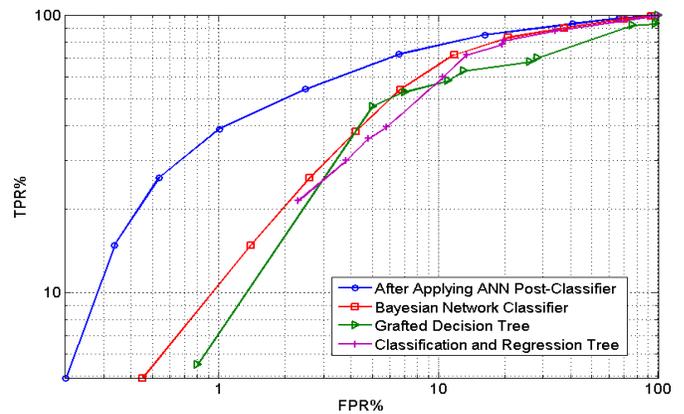

Figure 4. ROC of various common bioacoustic signal classifiers, and the effect of applying HK-ANN (the ANN Post-Classifier).

## 4. Discussion and Conclusion

The work herein considered a novel method, combining human intuition with an ANN classification stage for processing large amounts of passive acoustic data.

Since marine mammals typically exhibit seasonal patterns of occurrence and distribution, the automated algorithms are also expected to provide trends, as shown by a diel plot graphic. However, errors due to ambient noise and other conflicting acoustics events can pose significant challenges for automated algorithms, especially with larger datasets. Often times, developers do not have access to large amounts of data when developing recognition tools. Furthermore, background noise and other conditions can color recordings, offering bias and making pre-trained recognition algorithms prone to high error rates when running on large scale datasets. Results show that in highly noisy environments, training a basic classifier using a fixed feature set was not sufficient for building an effective automated classification stage for studying seasonal patterns for minke song activity. For this situation, excessive numbers of false positives destroys the basic seasonal migration pattern in the diel graph.

The proposed approach, referred to as the HK-ANN, was used to augment an ANN with a post-classifier stage by incorporating human knowledge. Using human scoring measure, along with the same feature set, provided a significant improvement in the automatic detection and classification of the signals of interest. Based on the results for the seasonal patterns, the post processing stage properly recognized minke whale songs and provided a seasonal pattern as shown in the diel plots.